%% file: acl_latex.tex
\pdfoutput=1

\documentclass[11pt]{article}

\usepackage[svgnames]{xcolor}
\usepackage{amsmath}
\usepackage{multirow}
\usepackage{booktabs}
\usepackage{tabularx}
\usepackage{tcolorbox}
\usepackage{subcaption}
\usepackage{listings}
\usepackage{algorithm}
\usepackage{algpseudocode}

\usepackage[preprint]{acl}

\usepackage{times}
\usepackage{latexsym}

\usepackage[T1]{fontenc}

\usepackage[utf8]{inputenc}

\usepackage{microtype}

\usepackage{inconsolata}

\usepackage{graphicx}
\usepackage{xspace}
\tcbuselibrary{breakable}
\newcommand{\CritiQ}[0]{\textsc{CritiQ}\xspace}
\title{\CritiQ: Mining Data Quality Criteria from Human Preferences}

\author{
\textbf{Honglin Guo\textsuperscript{1,2}},
\textbf{Kai Lv\textsuperscript{1,2}},
\textbf{Qipeng Guo\textsuperscript{2}\thanks{Corresponding authors.}},
\\
\textbf{Tianyi Liang\textsuperscript{3,2}},
\textbf{Zhiheng Xi\textsuperscript{1}},
\textbf{Demin Song\textsuperscript{2}},
\textbf{Qiuyinzhe Zhang\textsuperscript{4,2}},
\textbf{Yu Sun\textsuperscript{2}},
\\
\textbf{Kai Chen\textsuperscript{2}},
\textbf{Xipeng Qiu\textsuperscript{1}},
\textbf{Tao Gui\textsuperscript{1}$^*$}
\\
\textsuperscript{1}Fudan University,
\textsuperscript{2}Shanghai AI Laboratory,
\\
\textsuperscript{3}East China Normal University,
\textsuperscript{4}University of Science and Technology of China
\\
\texttt{hlguo24@m.fudan.edu.cn}, \texttt{guoqipeng@pjlab.org.cn}, \texttt{tgui@fudan.edu.cn}
\\
\url{https://github.com/KYLN24/CritiQ}
}

\begin{document}
  \maketitle
  \begin{abstract}
    \input{00abstract}
  \end{abstract}

  \section{Introduction}
  \input{01introduction}

  \section{Related Work}
  \input{02relatedwork}

  \section{Method}
  \input{03method}

  \section{Experiments}
  \label{sec:experiments}
  \input{04experiments}

  \section{Analysis}
  \input{05analysis}

  \section{Conclusion}
  \input{06conclusion}

  \section*{Limitations}
  Our work has several limitations. First, our experiments focus on three
  specific domains, leaving the question of general domain data selection unexplored.
  The challenge of guiding annotators to provide quality comparisons in general
  domains remains open. Furthermore, while deriving criteria directly from human-annotated
  pairwise comparisons reduces biases compared to handwritten criteria, human biases
  can not be completely eliminated from the annotation process, as defining high-quality
  data remains inherently subjective. Finally, due to computational constraints,
  we limit our approach to continual pretraining rather than pretraining from scratch,
  and use a relatively modest model with 3B parameters. Future work could explore scaling
  to larger models and more comprehensive training approaches.

  \bibliography{custom}

  \newpage
  \appendix
  \input{appendix}
\end{document}

%% file: 00abstract.tex
Language model heavily depends on high-quality data for
optimal performance. Existing approaches rely on manually designed heuristics, the perplexity of existing models, training classifiers, or
careful prompt engineering, which require significant expert experience and human annotation effort while
introduce biases. We introduce \textbf{\CritiQ}~\footnote{\textbf{Crit}er\textbf{i}a
of Data \textbf{Q}uality, pronounced as ``critic''.}, a novel data selection method that
automatically mines criteria from human preferences for data quality with only $\sim$30 human-annotated
pairs and performs efficient data selection. The main component, \CritiQ Flow, employs a manager agent to evolve
quality criteria and worker agents to make pairwise judgments. We build a knowledge
base that extracts quality criteria from previous work to boost \CritiQ Flow. Compared to perplexity- and classifier-based methods, verbal criteria are more interpretable and have greater reusable value. After deriving the criteria, we train the \CritiQ Scorer to give quality scores and
perform efficient data selection. We demonstrate the effectiveness of our method
in the code, math, and logic domains, achieving high accuracy on human-annotated test
sets. To validate the quality of the selected data, we continually train
\texttt{Llama 3.2} models and observe improved performance on downstream tasks compared
to uniform sampling. Ablation studies validate the benefits of the knowledge
base and the reflection process. We analyze how criteria evolve and the effectiveness
of majority voting.

%% file: 01introduction.tex
Large language models (LLMs) show significant performance in various downstream
tasks~\citep{brown_language_2020,openai_gpt-4_2024,dubey_llama_2024}. Studies
have found that training on high quality corpus improves the ability of LLMs
to solve different problems such as writing code, doing math exercises, and
answering logic questions~\citep{cai_internlm2_2024,deepseek-ai_deepseek-v3_2024,qwen_qwen25_2024}.
Therefore, effectively selecting high-quality text data is an important subject for
training LLM.

\begin{figure}[t]
    \centering
    \includegraphics[width=\linewidth]{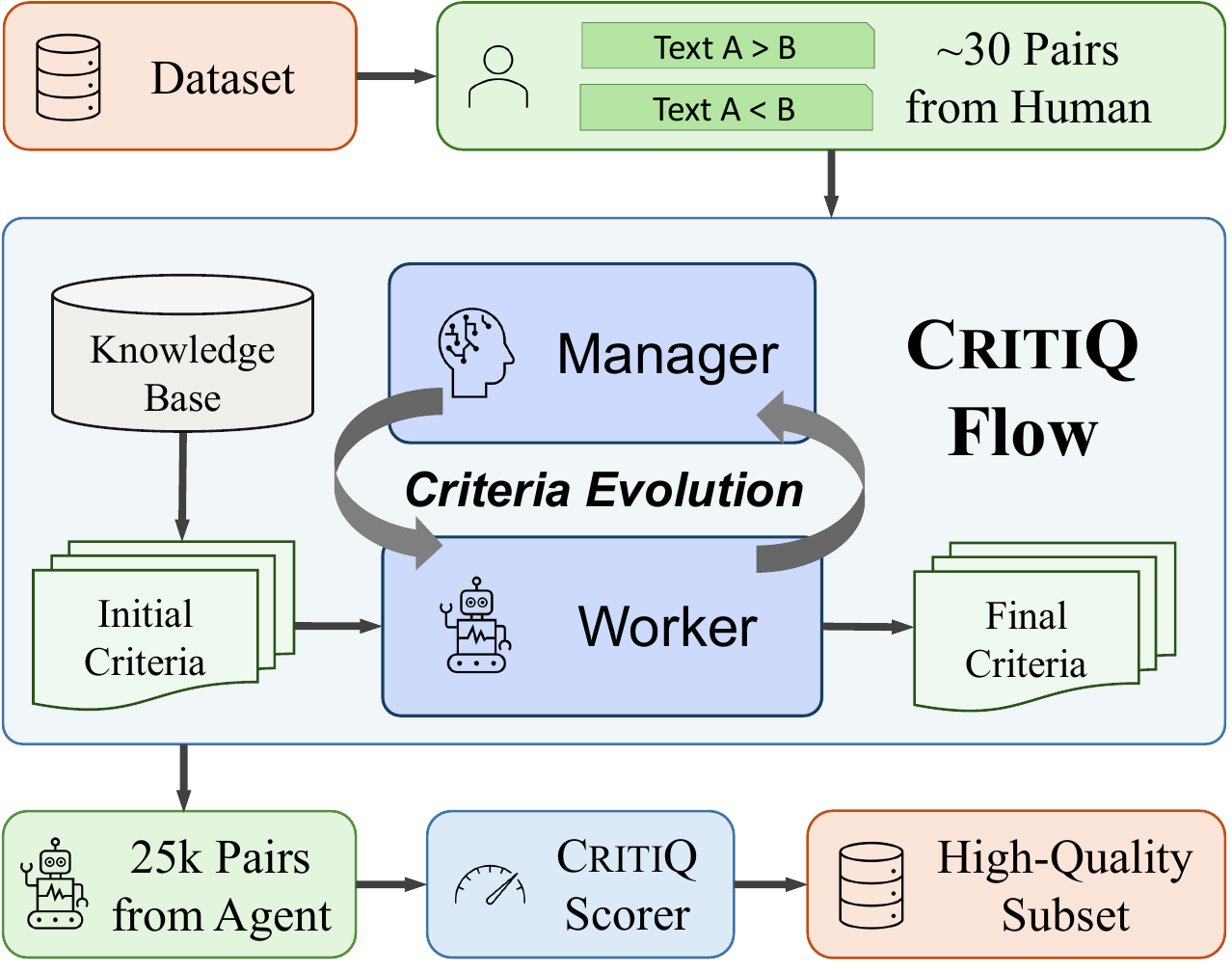}
    \caption{The overview of \CritiQ. We (1) employ human annotators to annotate $\sim$30
    pairwise quality comparisons, (2) use \CritiQ Flow to mine quality criteria, (3)
    use the derived criteria to annotate 25k pairs, and (4) train the \CritiQ Scorer to
    perform efficient data selection.}
    \label{fig:overview}
\end{figure}

To select high-quality data from a large corpus, researchers manually design heuristics~\citep{dubey_llama_2024,rae_scaling_2022},
calculate perplexity using existing LLMs~\citep{marion2023moreinvestigatingdatapruning,wenzek2019ccnetextractinghighquality},
train classifiers~\citep{brown_language_2020,dubey_llama_2024,xie_data_2023} and
query LLMs for text quality through careful prompt engineering~\citep{gunasekar_textbooks_2023,wettig_qurating_2024,sachdeva_how_2024}.
Large-scale human annotation and prompt engineering require a lot of human
effort. Giving a comprehensive description of what high-quality data is like is also
challenging. As a result, manually designing heuristics lacks robustness and introduces
biases to the data processing pipeline, potentially harming model performance
and generalization. In addition, quality standards vary across different
domains. These methods can not be directly applied to other domains without significant
modifications.

To address these problems, we introduce \CritiQ, a novel method to automatically
and effectively capture human preferences for data quality and perform efficient data
selection. Figure~\ref{fig:overview} gives an overview of \CritiQ, comprising an agent
workflow, \CritiQ Flow, and a scoring model, \CritiQ Scorer. Instead of manually describing
how high quality is defined, we employ LLM-based agents to summarize quality
criteria from only $\sim$30 human-annotated pairs.

\CritiQ Flow starts from a knowledge base of data quality criteria. The worker
agents are responsible to perform pairwise judgment under a given
criterion. The manager agent generates new criteria and refines them through reflection
on worker agents' performance. The final judgment is made by majority voting among
all worker agents, which gives a multi-perspective view of data quality.

To perform efficient data selection, we employ the worker agents to annotate a randomly
selected pairwise subset, which is ~1000x larger than the human-annotated one.
Following \citet{korbak_pretraining_2023,wettig_qurating_2024}, we train \CritiQ
Scorer, a lightweight Bradley-Terry model~\citep{bradley_rank_1952} to convert
pairwise preferences into numerical scores for each text. We use \CritiQ Scorer to
score the entire corpus and sample the high-quality subset.

For our experiments, we established human-annotated test sets to quantitatively
evaluate the agreement rate with human annotators on data quality preferences. We implemented the manager agent by \texttt{GPT-4o} and the worker
agent by \texttt{Qwen2.5-72B-Insruct}. We conducted experiments on different
domains including code, math, and logic, in which \CritiQ Flow showed a consistent
improvement in the accuracies on the test sets, demonstrating the effectiveness
of our method in capturing human preferences for data quality. To validate the quality
of the selected dataset, we continually trained \texttt{Llama 3.2}~\citep{dubey_llama_2024}
models and found that the models achieve better performance on downstream tasks
compared to models trained on the uniformly sampled subsets.

We highlight our contributions as follows. We will release the code to facilitate
future research.

\begin{itemize}
    \item We introduce \CritiQ, a method that captures human preferences for data
        quality and performs efficient data selection at little cost of human
        annotation effort.

    \item Continual pretraining experiments show improved model performance in code,
        math, and logic tasks trained on our selected high-quality subset compared to the raw dataset.

    \item Ablation studies demonstrate the effectiveness of the knowledge base and
        the the reflection process.
\end{itemize}

\begin{figure*}[t]
    \centering
    \includegraphics[width=\linewidth]{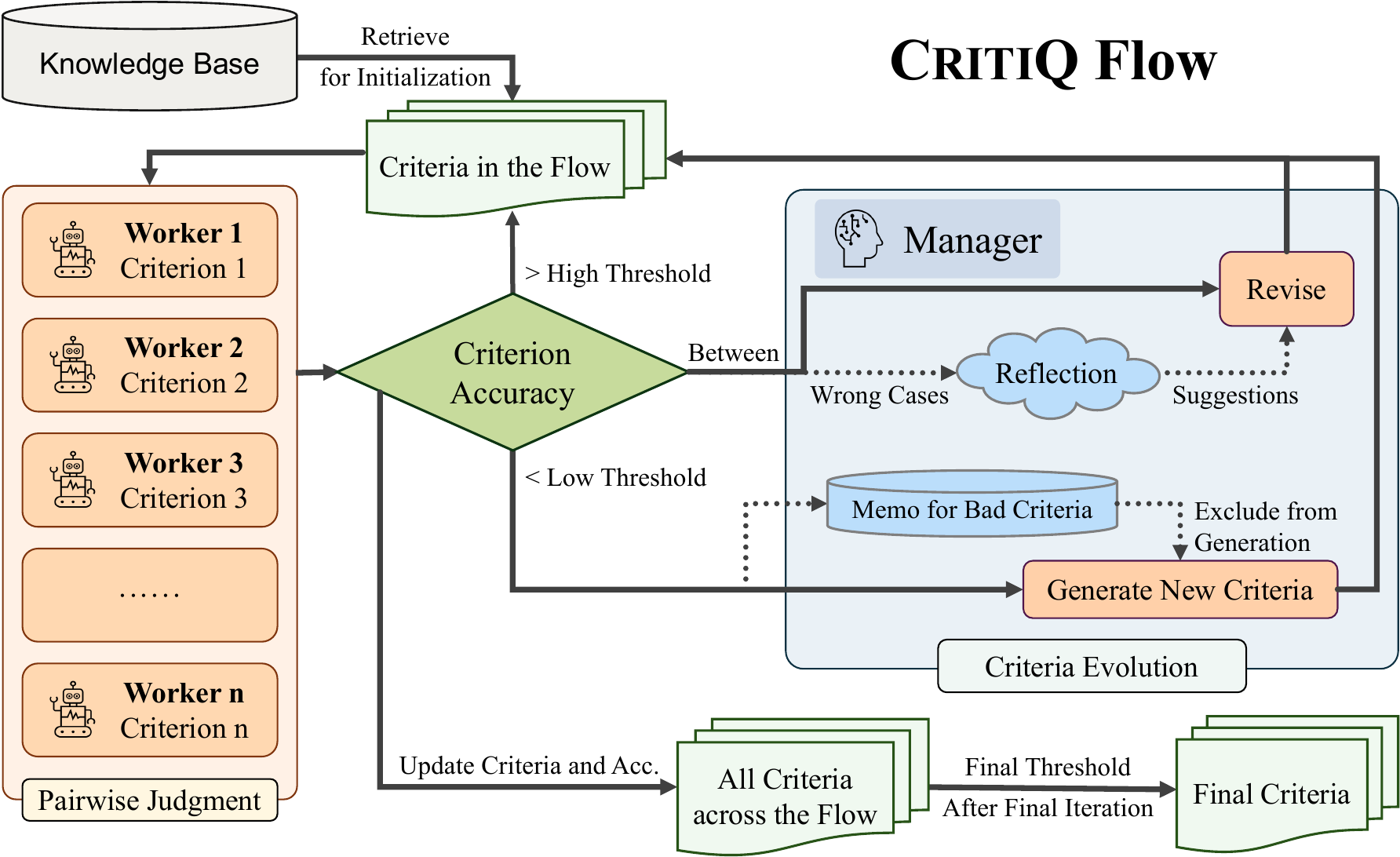}
    \caption{\CritiQ Flow comprises two major components: multi-criteria pairwise
    judgment and the criteria evolution process. The multi-criteria pairwise
    judgment process employs a series of worker agents to make quality
    comparisons under a certain criterion. The criteria evolution process aims to
    obtain data quality criteria that highly align with human judgment through
    an iterative evolution. The initial criteria are retrieved from the
    knowledge base. After evolution, we select the final criteria to annotate
    the dataset for training \CritiQ Scorer.}
    \label{fig:method}
\end{figure*}

%% file: 02relatedwork.tex
\paragraph{Heuristics for Data Selection.}

Using manually designed heuristics to identify data with specific
characteristics is a basic approach for data selection. Common rules include keyword
or stopword matching, length-based filtering, data source filtering, in-document
duplication~\citep{dubey_llama_2024,cai_internlm2_2024}, and training
classifiers~\citep{noauthor_improving_2024,xie_data_2023, dubey_llama_2024, wei_arctic-snowcoder_2024,korbak_pretraining_2023,lv_longwanjuan_2024}.
Designing these rules requires much experience and human effort.

Researchers also design specific rules to select high-quality domain data~\citep{wang_generative_2023,song2024code,lozhkov_starcoder_2024,huang_opencoder_2024},
which requires much expert experience and lacks scalability and generalization.

\paragraph{Quality Signals from LLMs.}

The use of LLMs to assess data quality has become a prevalent approach. Researchers
employ manual-designed prompts to query LLMs for quality assessment~\citep{dubey_llama_2024, sachdeva_how_2024,zhang_autonomous_2024},
often using educational value as a proxy for data quality~\citep{gunasekar_textbooks_2023, wei_arctic-snowcoder_2024}.
However, their focus remains limited to fixed aspects of data quality. Although these
methods reduce the need for human annotation, they introduce inherent biases
through predefined rules and standards.

Previous works like QuRating~\citep{wettig_qurating_2024} evaluate data quality using
multiple manually defined criteria including writing style, factual accuracy, level
of expertise, and educational value. These predefined criteria show varying effectiveness across
different domains, suggesting that manually summarized criteria lack generalization
and can not accurately describe data quality. In contrast, \CritiQ Flow automatically
discovers quality criteria by effectively capturing human preferences about
data quality assessment from a small number of human-annotated pairs.

\paragraph{Thought and Reflection of LLMs.}

Prompting
LLMs to reason before giving the final answer improves the model's
performance on various tasks~\citep{kojima_large_2023,yao_react_2023}. In our work, we also require the agents to think
and analyze before making the quality comparison.

Reflection is a common technique to improve the performance of LLMs through
iterative critiquing and refinement~\citep{shinn_reflexion_2023,madaan_self-refine_2023,saunders_self-critiquing_2022,xi2024enhancingllmreasoningcritique}.
Existing frameworks have integrated the reflection mechanism to build LLM-based
agents and do prompt engineering~\citep{yuksekgonul_textgrad_2024,asai_self-rag_2023,wu_autogen_2023}.
In \CritiQ Flow, we also prompt the agent to
examine the wrong predictions and refine the quality criteria accordingly.

%% file: 03method.tex
\subsection{Overview}

In \CritiQ, we first use an agent workflow, \CritiQ Flow, to automatically extract
quality criteria from human preferences for data quality with limited human
annotation, and then use these criteria to train a scoring model, \CritiQ Scorer,
to efficiently perform large-scale data selection.

For a specific text dataset $D$, we sample $\sim$30 pairs of data points. In contrast to existing approaches that require the researchers to manually design prompts~\citep{dubey_llama_2024, sachdeva_how_2024,zhang_autonomous_2024,gunasekar_textbooks_2023, wei_arctic-snowcoder_2024},
a small amount of data annotation in our work requires less human effort. We employ human expert
annotators to determine which data point in each pair is of higher quality,
forming the training set $D_{\text{human}}$ for \CritiQ Flow. Details for annotation
are shown in Appendix~\ref{sec:annotation}. Figure~\ref{fig:method} shows how
\CritiQ Flow mines quality criteria from $D_{\text{human}}$. Prompts we used are shown
in Appendix~\ref{sec:appendix_prompts}.

To perform large-scale data selection, we train \CritiQ Scorer, a lightweight scoring
model. Following~\citet{korbak_pretraining_2023,wettig_qurating_2024}, we use a
Bradley-Terry model~\citep{bradley_rank_1952} to convert the pairwise comparison
into a numerical score. We randomly sample a larger number of text pairs, forming
the training dataset $D_{\text{agent}}$ for \CritiQ Scorer. The quality
preference labels will be annotated by the worker agents through the pairwise
judgment process under the obtained quality criteria. Finally, we use \CritiQ
Scorer to score all text data in $D$ and select the high-quality subset
according to the quality scores.

\begin{table*}
    [t]
    \centering
    \begin{tabular}{llcccccccc}
        \toprule \multicolumn{2}{l}{\textbf{Method}} & \textbf{Code}     & $\Delta$                                  & \textbf{Math}                    & $\Delta$                               & \textbf{Logic}                   & $\Delta$                               & \textbf{Avg.}                 & $\Delta$                               \\
        \midrule \multicolumn{2}{l}{Vanilla}         & 82.02             & -                                         & 72.86                            & -                                      & 72.99                            & -                                      & 75.96                         & -                                      \\
        \midrule \multicolumn{2}{l}{TextGrad}        & 72.70             & \textcolor{DarkRed}{\ \ \ -9.32}          & 78.57                            & \textcolor{DarkGreen}{\ \ +5.71}       & 75.22                            & \textcolor{DarkGreen}{\ \ +2.23}       & 75.50                         & \textcolor{DarkRed}{\ \ \ -0.46}       \\
        \multirow{4}{*}{\rotatebox{90}{QuRating}}    & Writing Style     & 73.03                                     & \textcolor{DarkRed}{\ \ \ -8.99} & 52.86                                  & \textcolor{DarkRed}{\ -20.00}    & 59.70                                  & \textcolor{DarkRed}{\ -13.29} & 61.86                                 & \textcolor{DarkRed}{\ -14.09}    \\
                                                     & Facts \& Trivia   & 76.40                                     & \textcolor{DarkRed}{\ \ \ -5.62} & 44.29                                  & \textcolor{DarkRed}{\ -28.57}    & 84.33                                  & \textcolor{DarkGreen}{+11.34} & 68.34                                 & \textcolor{DarkRed}{\ \ \ -7.62} \\
                                                     & Educational Value & 85.39                                     & \textcolor{DarkGreen}{\ \ +3.37} & 68.57                                  & \textcolor{DarkRed}{\ \ \ -4.29} & 84.33                                  & \textcolor{DarkGreen}{+11.34} & 79.43                                 & \textcolor{DarkGreen}{\ \ +3.47} \\
                                                     & Require Expertise & 79.21                                     & \textcolor{DarkRed}{\ \ \ -2.81} & 52.86                                  & \textcolor{DarkRed}{\ -20.00}    & 84.33                                  & \textcolor{DarkGreen}{+11.34} & 72.13                                 & \textcolor{DarkRed}{\ \ \ -3.82} \\
        \midrule \multicolumn{2}{l}{\CritiQ Flow}     & \textbf{89.33}    & \textcolor{DarkGreen}{\textbf{\ \ +7.31}} & \textbf{84.57}                   & \textcolor{DarkGreen}{\textbf{+11.71}} & \textbf{88.06}                   & \textcolor{DarkGreen}{\textbf{+15.07}} & \textbf{87.32}                & \textcolor{DarkGreen}{\textbf{+11.36}} \\
        \multicolumn{2}{l}{\quad w/o evo.}           & 86.40             & \textcolor{DarkGreen}{\ \ +4.38}          & 78.00                            & \textcolor{DarkGreen}{\ \ +5.14}       & 85.97                            & \textcolor{DarkGreen}{+12.98}          & 83.46                         & \textcolor{DarkGreen}{\ \ +7.50}       \\
        \multicolumn{2}{l}{\quad w/o k.b.}           & 87.19             & \textcolor{DarkGreen}{\ \ +5.17}          & 82.57                            & \textcolor{DarkGreen}{\ \ +9.71}       & 81.64                            & \textcolor{DarkGreen}{\ \ +8.65}       & 83.80                         & \textcolor{DarkGreen}{\ \ +7.84}       \\
        \multicolumn{2}{l}{\quad w/o evo. \& k.b.}   & 83.03             & \textcolor{DarkGreen}{\ \ +1.01}          & 76.29                            & \textcolor{DarkGreen}{\ \ +3.43}       & 68.36                            & \textcolor{DarkRed}{\ \ \ -4.63}       & 75.89                         & \textcolor{DarkRed}{\ \ \ -0.06}       \\
        \multicolumn{2}{l}{\CritiQ Scorer}            & \textbf{89.89}    & \textcolor{DarkGreen}{\textbf{\ \ +7.87}} & \textbf{90.00}                   & \textcolor{DarkGreen}{\textbf{+17.14}} & \textbf{90.22}                   & \textcolor{DarkGreen}{\textbf{+17.23}} & \textbf{90.04}                & \textcolor{DarkGreen}{\textbf{+14.08}} \\
        \bottomrule
    \end{tabular}
    \caption{Accuracies on the human-annotated $D_{\text{test}}$. The best results
    and the best results without training a model are in bold. ``$\Delta$'' is
    the delta value with the vanilla results. ``evo.'' for iterative criteria
    evolution. ``k.b.'' for retrieving initial criteria from the knowledge base
    instead of generating all initial criteria by the manager agent. The results
    are the average over 5 experiments with different random seeds.}
    \label{tab:main_results}
\end{table*}

\subsection{Knowledge Base}
\label{sec:knowledge_base}

As an iterative agent workflow, the quality of the initial criteria is crucial for \CritiQ Flow. Many research papers have shared valuable insights on quality standards and have succeeded in data selection. Therefore, we can leverage findings from these data selection studies to establish a criteria knowledge base. Drawing from well-validated methodologies, the knowledge base can enhance the initialization of \CritiQ Flow, ensuring a strong foundation for subsequent refinements.

To construct the knowledge base, we first crawl the cited papers of the datasets published on the Hugging Face
Hub~\footnote{\href{https://huggingface.co/datasets}{https://huggingface.co/datasets}
data collected before July 2024}. We
only use the arXiv papers available in HTML format, avoiding potential issues with PDF
parsing. We employ \texttt{GPT-4o-mini} to identify papers that introduce datasets
from the titles and abstracts. Subsequently, we use \texttt{GPT-4o-mini} to
systematically extract quality criteria from these papers. To perform deduplication, we normalize the name of each criterion and filter out those with a Jaccard similarity above the threshold of 0.3. Finally, we establish the knowledge base $C_{\text{knowledge}}$ comprising 342 distinct
quality criteria.

\begin{algorithm}
    [htbp]
    \caption{Retrieve Criteria from $C_{\text{domain}}$}
    \label{alg:kb}
    \begin{algorithmic}
        [1]
        \renewcommand{\algorithmicrequire}{\textbf{Input:}}
        \renewcommand{\algorithmicensure}{\textbf{Output:}}
        \Require $C_{\text{domain}}$, $D_{\text{human}}$, $n$ \Ensure $C_{\text{retrieved}}$
        \State Initialize $C_{\text{retrieved}}[\ ], \text{Acc}[\ ]$\For{$c_{i}\in C_{\text{domain}}$}
        \State $\text{Acc}_{i}\gets$acc over\ $ D_{\text{human}}$ \EndFor \State
        Sort $C_{\text{domain}}$ by $\text{Acc}$ \Comment{Descending order} \For{$c_{i}\in C_{\text{domain}}$}
        \If{$\Call{Length}{C_{\text{retrieved}}\geq n}$} \State break \EndIf \If{$\text{Acc}_{i}> 0.5$}
        \State \Call{Append}{$C_{\text{retrieved}}, c_{i}$} \EndIf \EndFor
        \State \Return $C_{\text{retrieved}}$
    \end{algorithmic}
\end{algorithm}

We use this $C_{\text{knowledge}}$ to provide initial criteria for \CritiQ Flow. We query
a model with the domain description of the dataset to retrieve potentially
useful criteria from $C_{\text{knowledge}}$, forming $C_{\text{domain}}$. As shown
in Algorithm~\ref{alg:kb}, we then retrieve $n$ criteria from $C_{\text{domain}}$.
If the criteria are not enough, we query the manager agent to propose new criteria.

\subsection{Multi-Criteria Pairwise Judgment}
\label{sec:voting}

Given a set of quality criteria $C$ and a pair of data points $p=(\text{text}_{\text{A}}
, \text{text}_{\text{B}}) \in D_{h}$, the pairwise judgment process gives a
quality preference by a worker agent. Each criterion has a corresponding description
to guide the comparison. In consideration of cost and efficiency, we do not use an
expensive model as the worker agent. Instead, we use a model that can perform simple
comparisons under a single criterion, which is not difficult for many open-source
LLMs.

For each criterion $c_{i}\in C$, we query a distinct worker agent to determine which
data point exhibits higher quality. The worker agent analyzes both data points
with respect to $c_{i}$ before making a judgment. If $c_{i}$ is not applicable or
if both text A and B of $p$ demonstrate comparable quality, the worker agent can
refuse to provide an answer, i.e., answer ``null''. The final judgment across all
criteria is made through majority voting, i.e.,
\[
    \text{judge}(p, C) = \text{majority}_{c_{i} \in C}(\{\text{worker}_{i}(p, c_{i}
    )\}),
\]
where $\text{worker}_{i}(p, c_{i}) \in \{\text{A}, \text{B}, \text{null}\}$ is
the worker agent's judgment of $p$ under $c_{i}$.

Because we only focus on whether the final judgment is consistent with the human
annotation and do not require all criteria to be applicable to a certain pair, we
do not take these situations into consideration when calculating the accuracy for
this criterion. The criterion accuracy for $c_{i}$ on dataset $D_{h}$ is
calculated as

\begin{footnotesize}
    \[
        \text{acc}(c_{i}|D_{h}) = \frac{|\{p \in D_{h}|\text{w}_{i}(p,c_{i})=\text{h}(p)\}|}{|D_{h}|-|\{p
        \in D_{h}|\text{w}_{i}(p,c_{i})=\text{null}\}|},
    \]
\end{footnotesize}

where $h(p) \in \{A, B\}$ is the human-annotated higher-quality one in $p$, and $\text{w}_{i}
(p,c_{i})$ is the worker agent's judgment of $p$ according to $c_{i}$.

\subsection{Criteria Evolution}

After retrieving the initial criteria from the knowledge base, we perform an iterative
criteria evolution to improve the accuracy on $D_{\text{human}}$. For each
iteration, we first make pairwise judgments on $D_{\text{human}}$. Based on the
accuracy $acc_{i}$ of each criterion $c_{i}$, we then divide them into three groups
by a high threshold $t_{\text{high}}$ and a low threshold $t_{\text{low}}$. For
$c_{i}$ with $acc_{i}\geq t_{\text{high}}$, we keep them directly. For $c_{i}$
with $acc_{i}\leq t_{\text{low}}$, we remove them and query the manager agent to generate
new criteria. Simultaneously, they will be recorded to avoid
being generated again by the manager agent in subsequent iterations. For $c_{i}$
with $t_{\text{low}}< acc_{i}< t_{\text{high}}$, we ask the manager agent to do reflection.
For each incorrect judgment of
$p\in \{p|worker(p, c_{i})\notin \{h(p ), null\} \}$, we provide the manager with
the right answer $h(p)$ and worker agent's thought. The manager agent should
analyze why the worker agent makes mistakes and provide a suggestion to itself
on how to improve the criteria. Given all suggestions from the wrong cases, the
manager agent should refine the description of $c_{i}$ as $c'_{i}$. $acc'_{i}$ will
be calculated in the next iteration.

Unlike the gradient descent algorithm, text-based optimization does not
guarantee that the loss will decrease within a neighborhood of the current state.
Therefore, we need to introduce external constraints to ensure this. In \CritiQ Flow,
we save all criteria $c_{i}$ throughout the evolution process with their
accuracies $acc_{i}$. After getting the new accuracy $acc'_{i}$ of a revised criterion
$c'_{i}$, we will only update the description of it when
$acc'_{i}\geq acc_{i}$. This constraint ensures that the description revision will
not make the criterion worse. The final criteria are those with the highest
accuracy of all criteria across iterations.

\subsection{Train the Scoring Model}

After obtaining the quality criteria, we can use them to annotate a larger
number of pairs from the dataset $D$ to train \CritiQ Scorer. To form the pairs,
we randomly sample several data points and group them by the length of the text
to remove the potential influences of length biases of the worker
agent. We then use the pairwise judgment process to annotate the pairs according to the
quality criteria mined by \CritiQ Flow, forming $D_{\text{agent}}$. Only worker agents are employed in
this process, which get rid of the high cost API calls to the manager agent.

Training the \CritiQ Scorer $s_{\theta}$ is straightforward by minimizing the loss
function,
\[
    \mathcal{L}(\theta) = - \frac{1}{N}\sum_{p \in D_{\text{agent}}}\log\sigma (s
    _{\theta}(d_{\text{high}}) - s_{\theta}(d_{\text{low}})),
\]
where $\sigma$ is the sigmoid function, $d_{\text{high}}$ and $d_{\text{low}}$ are
the relatively high- and low- quality data points in the pair $p$.

\subsection{Selecting Data}

In consideration of cost and efficiency, we use a lightweight base model as the scoring
model, which increases the speed of scoring the entire dataset $D$. After getting
a score $s_{\theta}(d_{i})$ for each data point $d_{i}$ in $D$, we normalize the
scores to obtain the final quality score $s_{i}$. As QuRating~\citep{wettig_qurating_2024}
suggests, sampling is better than naive top-$k$ selection. We select each data
point $d_{i}$ with the probability $p_{i}\propto exp(\frac{s_{i}}{\tau})$, where $\tau$
is the temperature. This process is implicitly equivalent to reward-weighted
regression~\citep{wettig_qurating_2024,korbak_pretraining_2023,peters_reinforcement_2007}.
We use the Gumbel top-$k$ trick~\citep{wettig_qurating_2024,kool_stochastic_2019}
to perform efficient sampling without replacement.

%% file: 04experiments.tex
We verify the effectiveness of \CritiQ Flow in improving the accuracies on human-annotated
test sets. Hyperparameters for \CritiQ Flow are
shown in Appendix~\ref{sec:hyperparameter}. We continually pretrain a \texttt{Llama-3.2-3B}
model to show the improved quality of our selected subset compared to the
original dataset.

\subsection{Setup}

\paragraph{Dataset.}

We focus on three domains: code, math and logic. We use the Python subset of the
Stack v2~\citep{lozhkov_starcoder_2024}, the non-code subset of OpenWebMath~\citep{paster_openwebmath_2023}
and Zyda-2~\citep{tokpanov_zyda-2_2024} datasets as the source dataset $D$. The
numbers of pairs of $D_{\text{human}}$ and $D_{\text{agent}}$ are shown in Table~\ref{tab:number_dataset_pairs}.

\begin{table}[htbp]
    \centering
    \footnotesize
    \begin{tabular}{lccc}
        \toprule \textbf{Domain} & \#$D_{\text{human}}$ & \#$D_{\text{agent}}$ & \#$D_{\text{test}}$ \\
        \midrule \textbf{Code}   & 40                   & 25000                & 178                 \\
        \textbf{Math}            & 30                   & 25000                & 70                  \\
        \textbf{Logic}           & 30                   & 25000                & 134                 \\
        \bottomrule
    \end{tabular}
    \caption{Number of pairs in each split.}
    \label{tab:number_dataset_pairs}
\end{table}

\paragraph{Models.}
We employ \texttt{GPT-4o}~\footnote{The specific version is \texttt{gpt-4o-2024-11-20}.}
as the manager agent which is good at reflection but is costly, and \texttt{Qwen2.5-72B-Instruct}
as the worker agent which can perform simple pairwise comparison while is
relatively cheap. We initialize \CritiQ Scorer by \texttt{Qwen2.5-1.5B} for
efficiency considerations. Hyperparameters for \CritiQ Scorer are shown in Appendix~\ref{sec:hyperparameter}.

\paragraph{Baselines.}
Directly prompting the worker LLM for data quality comparison serves a vanilla
baseline. We use the same prompt as ours without specifying a criterion for vanilla
baseline experiments. We compare the optimization algorithm in our workflow with
TextGrad~\citep{yuksekgonul_textgrad_2024}. The initial prompt for TextGrad is
the same as the vanilla baseline. We run TextGrad optimizations on the same training
set $D_{\text{agent}}$ as ours. We compare our criteria with those proposed by
QuRating~\citep{wettig_qurating_2024}. The prompts for QuRating are from their
original work.

\paragraph{Evaluation.}
We evaluate \CritiQ Flow by the accuracy on the human-annotated test set $D_{\text{test}}$.
High accuracy indicates effectiveness in capturing human preferences for data
quality. For each pair, three annotators will determine which data point
exhibits higher quality independently under the same annotation guidelines with
$D_{\text{human}}$. We only keep the pairs for which all three annotators give the
same judgment. The final number of pairs in $D_{\text{test}}$ is shown in Table~\ref{tab:number_dataset_pairs}.
We emphasize that although we take human effort to annotate more pairs for validation
purpose, and the workflow itself just needs a tiny annotated dataset to work. We will
show how well \CritiQ Flow mines data quality criteria by only $\sim$30 human
annotated pairs and gets high accuracies on $D_{\text{test}}$.

\subsection{Results}

We report the accuracies of the baselines and \CritiQ on the test set of all 3 domains
in Table~\ref{tab:main_results}. In addition, we report the ablation results for
the knowledge base and the criteria evolution process.

\paragraph{Vanilla method can be improved by TextGrad and \CritiQ Flow.}
Although the vanilla method is not low in the agreement rate with human
annotators, it can be further improved by TextGrad~\citep{yuksekgonul_textgrad_2024}
and \CritiQ Flow. Detailed descriptions and instructions help the worker agent to
perform better judgments.

\paragraph{\CritiQ Flow outperforms TextGrad.}
Compared with TextGrad, \CritiQ Flow achieves higher accuracies in all domains,
indicating higher effectiveness in capturing human preferences for data quality.
Interestingly, we find that TextGrad is also trying to find quality criteria, but
it is not as effective as \CritiQ Flow. This suggests that the optimization
algorithm in our workflow is more effective in the scenarios of mining quality
criteria from human preferences. We show the prompts generated by TextGrad in Appendix~\ref{sec:textgrad}.

\paragraph{\CritiQ Flow surpasses single criteria.}
Any single criterion proposed by QuRating~\citep{wettig_qurating_2024} fails to achieve
a high accuracy. Although, as highlighted in many related studies~\citep{zhang_autonomous_2024,gunasekar_textbooks_2023,wei_arctic-snowcoder_2024},
the Educational Value criterion shows relatively higher consistency with human judgment,
it can not comprehensively describe data quality. This suggests that compared to
single criterion, and \CritiQ Flow which uses multiple criteria is better.

\paragraph{Evolution and knowledge base help \CritiQ Flow improve the
performance.}
Ablation shows that both the iterative evolution process and knowledge base in our
workflow help improve the accuracies. This indicates that the criteria extracted
from previous work are effective in judging data quality, while still have the potential
to be optimized according to the specific domain and dataset; and that the
optimization process is effective in improving the criteria with only $\sim$30 human annotations.

\paragraph{\CritiQ Scorer shows increased accuracy.}
Notably, \CritiQ Scorer achieves higher accuracies than the direct multi-criteria
voting by worker agents across all domains, despite being trained on data
annotated by them. This suggests that our method effectively extracts human's inner
quality evaluation criteria, and these criteria demonstrate strong generalization
capability.

\subsection{Continual Pretraining}

We choose \texttt{Llama-3.2-3B} as the base model for the continual pretraining experiments.
We sample 10B tokens from the Stack v2 and Zyda-2, and 3B from OpenWebMath. We
perform uniform sampling and sampling using \CritiQ Scorer with temperature
$\tau=1$ for the code and math datasets and $\tau=0.5$ for the logic dataset. For comparison, we use QuRator~\citep{wettig_qurating_2024} to sample the subsets under the ``educational value'' criterion, which is the criterion showing the highest accuracy in Table~\ref{tab:main_results}.
We continually train the models on the nine datasets separately. Hyperparameters
are shown in Appendix~\ref{sec:hyperparameter}.

We evaluate the continually trained models on corresponding downstream tasks, including
4 code-writing tasks: HumanEval~\citep{chen_evaluating_2021}, MBPP~\citep{austin_program_2021},
HumanEval+, and MBPP+~\citep{liu_is_2023}; 3 math problem solving tasks: GSM8k~\citep{cobbe_training_2021},
SAT-Math~\citep{zhong_agieval_2023}, and MATH~\citep{hendrycks_measuring_2021};
and 2 logic reasoning tasks ARC-Challenge~\citep{clark_think_2018} and LogiQA~\citep{zhong_agieval_2023}.
Coding tasks are evaluated using EvalPlus~\citep{liu_is_2023}, while others are evaluated
by OpenCompass~\citep{2023opencompass}. For MATH~\citep{hendrycks_measuring_2021}, we used the AGIEval~\citep{zhong_agieval_2023} sampled subset. The results are shown in Table~\ref{tab:continual_pretraining}.
The models trained on our selected high-quality subsets show improved
performance on downstream tasks compared to the models trained on the uniformly
sampled subsets.

\begin{table}[htbp]
    \centering
    \footnotesize
        \begin{tabularx}
            {\linewidth} {l@{\hskip 8pt}c@{\hskip 8pt}c@{\hskip 8pt}c}
            \toprule
            \textbf{Code}
            & \textbf{HumanEval / +} & \textbf{MBPP / +} & \textbf{Avg. / +} \\
            \midrule
            Raw & 28.66 / 25.61 & 48.94 / 39.15 & 38.80 / 32.38 \\
            Stack & 31.71 / 27.44 & 56.61 / 46.30 & 44.16 / 36.87 \\
            QR-Edu & 36.59 / 32.32 & 59.26 / 47.35 & 47.93 / 39.84 \\
            \CritiQ & \textbf{39.02} / \textbf{33.54}
            & \textbf{68.73} / \textbf{48.41} & \textbf{53.88} / \textbf{40.98} \\
        \end{tabularx}
        \begin{tabularx}
            {\linewidth} {l@{\hskip 15pt}c@{\hskip 15pt}c@{\hskip 15pt}c@{\hskip 15pt}c}
            \midrule
            \textbf{Math} & \textbf{GSM8k} & \textbf{SAT-Math} & \textbf{MATH} & \textbf{Avg.} \\
            \midrule
            Raw & 27.60 & 35.00 & 5.50 & 22.70 \\
            OWM & 28.51 & 32.27 & 5.80 & 22.19 \\
            QR-Edu & 23.50 & 36.36 & 6.20 & 22.02\\
            \CritiQ & \textbf{32.22} & \textbf{39.55} & \textbf{6.34} & \textbf{26.04} \\
        \end{tabularx}
        \begin{tabularx}
            {\linewidth} {l@{\hskip 32.3pt}c@{\hskip 32.3pt}c@{\hskip 32.3pt}c}
            \midrule
            \textbf{Logic} & \textbf{ARC-C} & \textbf{LogiQA} & \textbf{Avg.} \\
            \midrule
            Raw & 37.97 & 27.34 & 32.66 \\
            Zyda-2 & 36.61 & 23.50 & 30.06 \\
            QR-Edu & 35.59 & 26.88 & 31.24 \\
            \CritiQ & \textbf{38.31} & \textbf{30.41} & \textbf{34.36} \\
            \bottomrule
        \end{tabularx}
    \caption{Evaluation results on downstream tasks of the continually trained
    model. ``Raw'' is the original \texttt{Llama-3.2-3B} model without any
    continual pretraining. ``+'' stands for HumanEval+ or MBPP+~\citep{liu_is_2023}, ``Stack''
    for the Python subset of the Stack v2~\citep{lozhkov_starcoder_2024}, and ``OWM'' for
    the non-code subset of OpenWebMath~\citep{paster_openwebmath_2023}. ``QR-Edu'' indicates the subsets sampled by QuRators~\citep{wettig_qurating_2024} according to the ``Educational Value'' criterion.}
    \label{tab:continual_pretraining}
\end{table}

%% file: 05analysis.tex
\subsection{Evolution of Criteria Distribution}

\begin{figure*}[t]
    \centering
    \begin{subfigure}
        [b]{0.48\textwidth}
        \includegraphics[width=\linewidth]{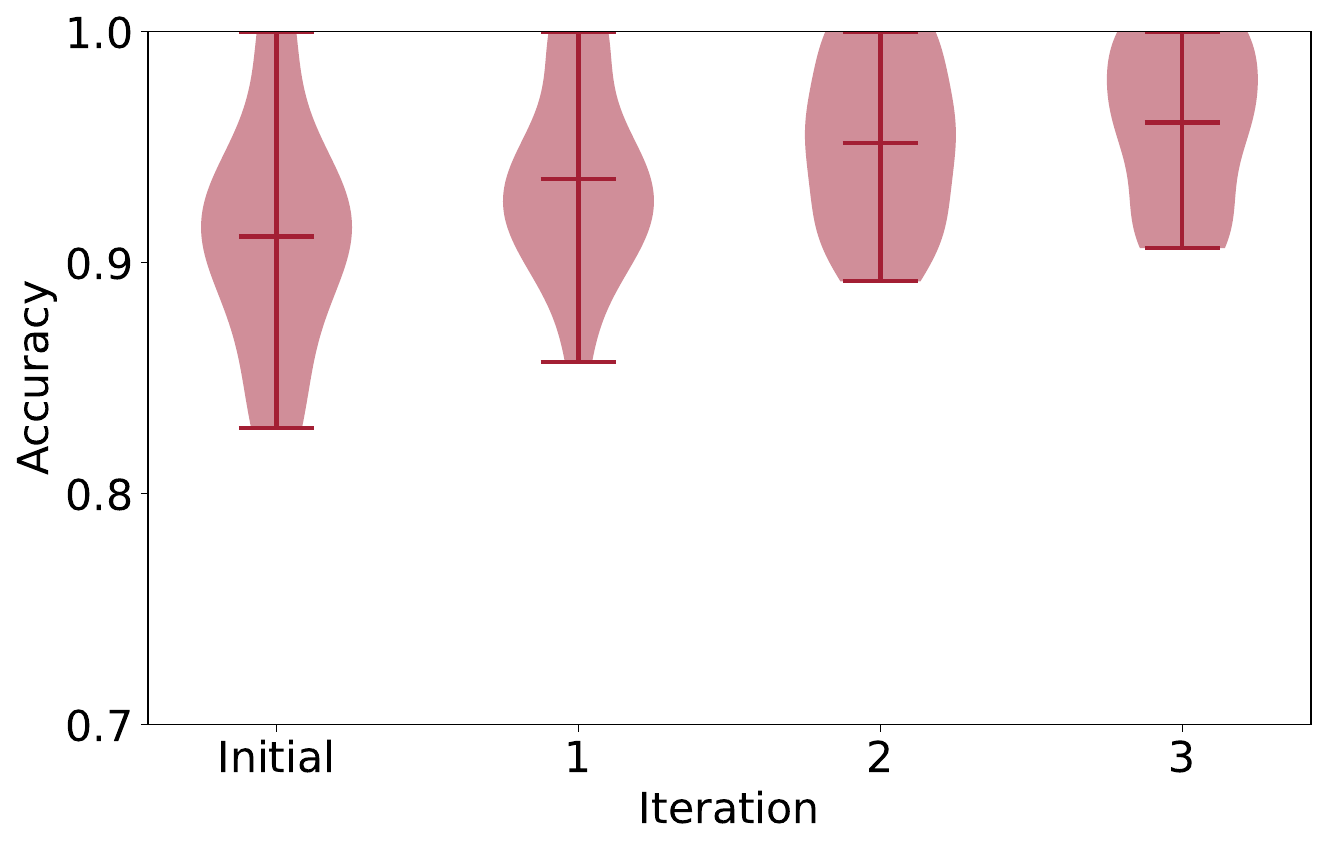}
        \caption{Distribution of accuracy.}
        \label{fig:evolution_acc}
    \end{subfigure}
    \hfill
    \begin{subfigure}
        [b]{0.48\textwidth}
        \includegraphics[width=\linewidth]{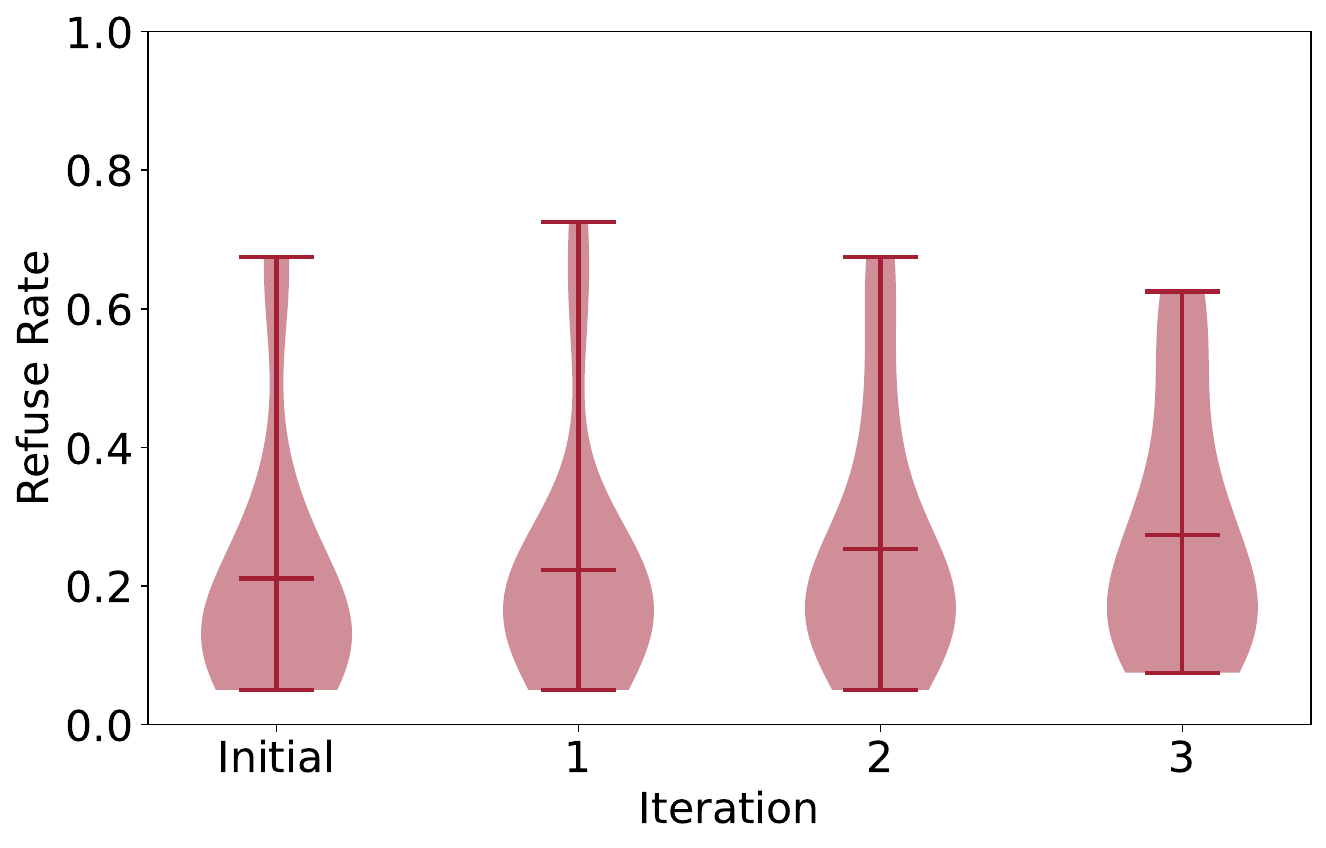}
        \caption{Distribution of refuse rate.}
        \label{fig:evolution_refuse_rate}
    \end{subfigure}
    \caption{Evolution of distributions of the top-$k$ Python code quality
    criteria through evolution iterations, where $k$ is the number of the final criteria.}
    \label{fig:criteria_evolution}
\end{figure*}

In this section, we analyze how the distribution of quality criteria evolves during
the evolution process. Using the code domain as a representative example, Figure~\ref{fig:evolution_acc}
shows the distribution of training accuracies for all criteria across optimization
iterations. The plot reveals a clear upward trend, with the distribution progressively
shifting and concentrating towards higher values as the optimization proceeds.
This trend demonstrates the effectiveness of our iterative optimization process.

Notably, several criteria achieve 100\% accuracy. As explained in Section~\ref{sec:voting},
we exclude the cases where the worker agent explicitly declines to provide a
judgment. Through the optimization process, the manager agent refines the
criteria descriptions to be more precise about their applicability. These highly
accurate criteria are particularly valuable as they effectively characterize code
quality and guide the worker agent to make accurate assessments when applicable,
even if they may not cover all possible scenarios.

In addition, we analyze the distribution of the refuse rate of the criteria. As
shown in Figure~\ref{fig:evolution_refuse_rate}, the refuse rate falls predominantly
in lower ranges, indicating that most criteria are widely applicable, while there
are still a few criteria with refuse rates higher than 60\% that are retained
due to their high accuracy when applicable.

\subsection{Criterion Refinement}
\label{sec:refinement}

The improvement in accuracy of \CritiQ Flow is driven by two key processes:
deprecating low-quality criteria and refining the mid-quality criteria by
revising the descriptions. Deprecating the low-quality ones is something like reject
sampling, which is straightforward in improving performance. In this section, we
analyze how mid-quality criteria are refined by the manager agent.

We categorize the criteria refinement into 2 types: (1) refining the criteria retrieved
from the knowledge base or generated by the manager agent, and (2) continually
refining the already refined criteria. We show examples of criteria before and after
refinement in Appendix~\ref{sec:appendix_ex_refinement}.

\paragraph{Refinement for Retrieved or Generated Criteria.}
The knowledge base is built on previous dataset research, so the criteria retrieved
from the knowledge base are often too general. When the knowledge base can not
provide enough criteria or some criteria are deprecated due to low accuracy, the
manager agent proposes new criteria. In this case, the initial descriptions of these
criteria are usually too vague, because they have not been evaluated by the
worker agent, thus the manager agent does not have enough information to generate
precise descriptions. As a result, the manager agent can refine those criteria by
rewriting them to fit the current domain, adding detailed guidelines for the
worker agent, and specifying the applicability.

\paragraph{Refinement for Refined Criteria.}
For previously refined criteria, the manager agent can further improve them by adding
more detailed descriptions or examples. However, we also observe that despite the
iterative optimization process, refinements do not always yield higher accuracy,
especially for already well-refined criteria. Excessive refinement by the
manager agent can lead to over-fitting, particularly with small training sets.
To address this, we encourage the manager agent to keep the criteria simple and concise.

\subsection{Majority Voting}

\begin{table}[htbp]
    \centering
    \footnotesize
    \begin{tabular}{lcccc}
        \toprule      & \textbf{Code} & \textbf{Math} & \textbf{Logic} & \textbf{Avg.} \\
        \midrule Ours & \textbf{89.33}         & \textbf{84.57}         & \textbf{88.06}          & \textbf{87.32}         \\
        w/o voting    & 84.16         & 81.14         & 85.22          & 83.51         \\
        \bottomrule
    \end{tabular}
    \caption{Accuracies with / without Majority Voting on the human-annotated
    $D_{\text{test}}$ across 3 domains. The higher values are in bold.}
    \label{tab:majority_voting}
\end{table}

We have demonstrated the majority voting mechanism in Section~\ref{sec:voting}. In
this section, we investigate the impact of the voting mechanism by evaluating
the accuracy of combining all criteria into a single prompt. We use the same quality
criteria derived by \CritiQ Flow and query the worker agent for judgments. The
accuracies are shown in Table~\ref{tab:majority_voting}. In all domains, the
accuracy decreases without the majority voting mechanism, indicating that the majority
voting mechanism is essential for the performance of \CritiQ Flow.

%% file: 06conclusion.tex
We introduce \CritiQ, a novel method that automatically mines quality criteria
from human preferences for data quality with limited human annotation and performs
efficient data selection. It uses an agent workflow, \CritiQ Flow, to
effectively summarize quality criteria from only $\sim$30 human-annotated pairs. \CritiQ Flow achieves high accuracies on human-annotated
test sets. Efficient data selection is performed by the lightweight \CritiQ Scorer.
We train models on our selected subset and observe increased performance on code,
math and logic domains, compared to baselines. We release the code of \CritiQ to facilitate future research.

%% file: appendix.tex
\section{Hyperparameters}
\label{sec:hyperparameter}

\subsection{Hyperparameters for \CritiQ Flow}

We have manually tried different sets of hyperparameters and the chosen
hyperparameters for the final experiments are shown in Table~\ref{tab:hyperparameter}.

\begin{table}[htbp]
    \centering
    \begin{tabular}{lccc}
        \toprule            & \textbf{Code} & \textbf{Math} & \textbf{Logic} \\
        \midrule \#Criteria & 20            & 20            & 20             \\
        \#Iterations        & 3             & 5             & 3              \\
        Retrieval Threshold & 0.5           & 0.5           & 0.5            \\
        High Threshold      & 0.9           & 0.8           & 0.8            \\
        Low Threshold       & 0.8           & 0.7           & 0.7            \\
        Final Threshold     & 0.9           & 0.7           & 0.8            \\
        \bottomrule
    \end{tabular}
    \caption{Hyperparameters for \CritiQ Flow.}
    \label{tab:hyperparameter}
\end{table}

\subsection{Hyperparameters for \CritiQ Scorer}

We use the \texttt{trl}~\citep{vonwerra2022trl} library to train \CritiQ Scorers. On the 3 domains, we train each \CritiQ Scorer using AdamW~\citep{loshchilov_decoupled_2019}
optimizer with learning rate $2 \times 10^{-5}$ and weight decay $0.01$ for $4$
epochs. The learning rate warmups in the first $20\%$ training steps and cosine decay
in the rest steps. We truncate the text longer than $32,768$ tokens. The global training
batch size is $128$. We randomly select $5\%$ from the \CritiQ Scorer training
set $D_{\text{agent}}$ as the validation set, and use the rest to train the scoring
model. We save the model every $50$ training steps and select the checkpoint
with the best validation accuracy as the final \CritiQ Scorer.

\subsection{Hyperparameters for Continual Pretraining}

We use AdamW~\citep{loshchilov_decoupled_2019} optimizer with the maximum
learning rate $1\times 10^{-4}$, the minimal learning rate $1\times 10^{-5}$,
and weight decay $0.01$ for $4$ epochs. The learning rate increases in the first
$5\%$ training steps, and cosine decays in the rest steps. The training sequence
length is $8192$ and global batch size is $4M$ tokens. Each model is trained on
$32$ NVIDIA H800 GPUs.

\section{Additional Experiments}
\label{sec:add_exp}

We performed additional experiments and analysis on the code domain to further inspect the performance of \CritiQ.

\subsection{More Baseline for Text-Based Optimization}

We used DRPO~\citep{DBLP:conf/emnlp/SinglaWLAHX24} to do pairwise juedgments in the code domain. Results are shown in Table~\ref{tab:add_exp_drpo}. The output of DRPO is shown in Appendix~\ref{sec:drpo}.

\begin{table}[htbp]
    \centering
    \begin{tabular}{lcc}
    \toprule
        \textbf{Method} & \textbf{Code} & \textbf{$\Delta$} \\
    \midrule
        Vanilla         & 82.02         & -                 \\
        DRPO            & 68.99         & -13.03            \\
        TextGrad        & 72.70         & -9.32             \\
        \CritiQ Flow    & 89.33         & +7.31             \\
        \CritiQ Scorer  & 89.89         & +7.87             \\
    \bottomrule
    \end{tabular}
    \caption{The result for DRPO compared with other methods in the code domain. This table serves as a supplement to Table~\ref{tab:main_results}.}
    \label{tab:add_exp_drpo}
\end{table}

\subsection{Impact of \#$D_{\text{human}}$}

We re-split the human-annotated dataset in the code domain. We sample a new $D_{\text{test}}$ with \#$D_{\text{test}}=168$. We show performances of \CritiQ Flow on the new $D_{\text{test}}$ with different $\#D_{\text{human}}$ in Table~\ref{tab:abl_n_d_human}.

\begin{table}[htbp]
    \centering
    \begin{tabular}{cc}
        \toprule
         $\#D_{\text{human}}$ & Accuracy \\
        \midrule 
        30 & 85.71 \\
        40 & 88.10 \\
        50 & 88.69 \\
        \bottomrule
    \end{tabular}
    \caption{Accuracies of \CritiQ Flow on the new $D_{\text{test}}$ with different $\#D_{\text{human}}$}
    \label{tab:abl_n_d_human}
\end{table}

\subsection{Token Usage}

We show statistics of the number of input and output tokens of the manager and worker agents when running \CritiQ Flow in the code domain in Table~\ref{tab:token_stats}.

\begin{table}[htbp]
    \centering
    \begin{tabular}{lcc}
        \toprule
        \textbf{Model}       & \textbf{Input \#Token} & \textbf{Output \#Token} \\ \midrule
        GPT-4o               & 435724               & 32292                 \\
        Qwen2.5 & 19279063           & 3762693             \\ \bottomrule
    \end{tabular}
    \caption{Token usages of the manager agent (GPT-4o) and worker agent (Qwen2.5-72B-Instruct) when running \CritiQ Flow in code domain. Hyperparameters are shown in Appendix~\ref{sec:hyperparameter}.}
    \label{tab:token_stats}
\end{table}

\subsection{Statistics on \#Criterion}

We report the statistics of the number of criteria selected/revised/removed at the end of the iteration in Table~\ref{tab:stats_criteria_final}.

\begin{table}[htbp]
    \centering
    \begin{tabular}{lccc}
        \toprule
        \textbf{Domain}              & Code  & Math  & Logic \\
        \midrule
        \textbf{\#Final Criteria}    & 15    & 30    & 20    \\
        \textbf{\#Appeared Criteria} & 32    & 33    & 20    \\
        \textbf{Revised Ratio (\%)}  & 18.75 & 15.83 & 16.25 \\
        \textbf{Removed Ratio (\%)}  & 7.50  & 1.67  & 1.25  \\
        \bottomrule
    \end{tabular}
    \caption{Statistics of the number of criteria selected/revised/removed at the end of the iteration of 3 domain.}
    \label{tab:stats_criteria_final}
\end{table}

\subsection{Accuracy Changes}

We show how the accuracies of all criteria in the code domain is changing across optimization iterations in Table~\ref{tab:acc_change}.

\begin{table}[htbp]
    \centering
    \resizebox{\linewidth}{!}{
    \begin{tabular}{lcccc}
        \toprule
        \textbf{Criterion}                   & \textbf{Init.} & \textbf{Iter. 1} & \textbf{Iter. 2} & \textbf{Iter. 3} \\ \midrule
        non\_empty\_lines                    & 94.44            & 94.44                & 94.87                & 97.30                \\
        error\_analysis                      & 88.57            & 90.62                & 90.62                & 90.62                \\
        test\_case\_generation               & 91.17            & 93.94                & 93.94                & 94.29                \\
        Program Validity                     & 91.17            & 96.67                & 96.67                & 100.00               \\
        Instructional Keywords               & 100.00           & 100.00               & 100.00               & 100.00               \\
        solution\_quality\_assessment        & 91.43            & 93.94                & 96.88                & 96.88                \\
        AutomatedProcessing                  & 93.75            & 93.75                & 96.15                & 96.15                \\
        code\_functionality                  & 100.00           & 100.00               & 100.00               & 100.00               \\
        execution\_success                   & 90.91            & 90.91                & 90.91                & 90.91                \\
        correctness\_patterns                & 87.88            & 100.00               & 100.00               & 100.00               \\
        execution\_test\_pass                & 83.33            & 91.67                & 91.67                & 91.67                \\
        realistic\_query\_input              & 93.94            & 93.94                & 93.94                & 94.12                \\
        instruction\_following               & 80.56            & 80.56                & 80.56                & 80.56                \\
        completeness\_of\_explanation        & 82.86            & 82.86                & 82.86                & 82.86                \\
        commented\_context                   & 85.29            & 89.19                & 89.19                & 89.19                \\
        length\_and\_complexity\_filter      & 91.67            & 91.67                & 96.29                & 96.43                \\
        compilability                        & 91.67            & 91.67                & 91.67                & 91.67                \\
        Error Identification                 & 75.00            & 75.00                & 75.00                & 75.00                \\
        currency                             & 74.29            & 74.29                & 74.29                & 74.29                \\
        noise\_reduction                     & 81.48            & 81.48                & 81.48                & 81.48                \\
        readability\_and\_structure          & 0.00             & 74.29                & 74.29                & 74.29                \\
        algorithm\_efficiency                & 0.00             & 85.71                & 100.00               & 100.00               \\
        security\_practices                  & 0.00             & 62.96                & 62.96                & 62.96                \\
        modularity\_and\_reusability         & 0.00             & 0.00                 & 84.62                & 84.62                \\
        scalability\_practices               & 0.00             & 0.00                 & 80.00                & 80.00                \\
        error\_logging\_and\_debugging       & 0.00             & 0.00                 & 80.95                & 82.33                \\
        input\_sanitization\_and\_validation & 0.00             & 0.00                 & 74.07                & 74.07                \\
        dependency\_management               & 0.00             & 0.00                 & 65.38                & 65.38                \\
        documentation\_coverage              & 0.00             & 0.00                 & 0.00                 & 73.53                \\
        test\_coverage                       & 0.00             & 0.00                 & 0.00                 & 81.25                \\
        runtime\_optimization                & 0.00             & 0.00                 & 0.00                 & 97.06                \\
        user\_friendly\_interface            & 0.00             & 0.00                 & 0.00                 & 62.07                \\ \bottomrule
    \end{tabular}
    }
    \caption{Changes of accuracies of all criteria in the code domain is changing across optimization iterations. 0.00 means the criterion is not yet proposed in this iteration. The value is the accuracy of the criterion defined in Section~\ref{sec:voting}.}
    \label{tab:acc_change}
\end{table}

\section{Annotation}
\label{sec:annotation}

\subsection{Annotators}

Our annotation team consists of three annotators for each domain (code, math, and
logic). The annotators are paper authors who meet the following qualifications:

\begin{itemize}
    \item Hold bachelor's or master's degrees

    \item Have multiple years of professional programming experience

    \item Possess foundational mathematical knowledge

    \item Demonstrate competency in logical reasoning
\end{itemize}

The annotators volunteered their time without additional compensation. As authors
of the paper, they had a vested interest in producing high-quality annotations, since
the annotation results directly impacted the experimental outcomes and overall research
quality. The inner agreement rate among the annotators is shown in Table~\ref{tab:annot_agree}.

\begin{table}[htbp]
    \centering
    \begin{tabular}{lc}
        \toprule
        \textbf{Domain} & \textbf{Agreement Rate (\%)} \\
        \midrule
        Code   & 87.20               \\
        Math   & 78.87               \\
        Logic  & 82.00               \\
        \bottomrule
    \end{tabular}
    \caption{The inner agreement rate among the annotators.}
    \label{tab:annot_agree}
\end{table}

\subsection{Annotation Guidelines}

\subsubsection{Annotation Guidelines for Code}

Please compare the two Python Code files and choose the one of higher quality.

Low-quality code often has the following characteristics:

\begin{itemize}
    \item The code is badly formatted or has syntax errors.

    \item The code consists solely of comments or package imports, which is non-informative.

    \item The code only consists of simple class or function definitions, which
        is hard to understand without other files.

    \item The code just defines meaningless variables while do not perform any operations.

    \item The code is too simple.

    \item The code contains too much hard-coded data or is a configuration or an
        entrypoint file to a larger project, which is not helpful in learning
        programming.
\end{itemize}

High-quality code often has the following characteristics:

\begin{itemize}
    \item The code is educational for code starters, which shows basic
        programming principles, design patterns, or data structures.

    \item The code is a solution to an algorithm problem, which is beneficial
        for learning algorithm.

    \item The code is well-structured with proper code comments, which leads to
        high readability and maintainability.

    \item The code shows clear purpose and can accurately solve certain kind of problems,
        while keeps extensible and flexible.

    \item The code has self-contained classes or functions that can be understood
        without other files, which shows high simplicity and reusability.
\end{itemize}

Choose the better one of A and B according to the above guidelines and your preferences
for code quality. If the two files are of similar level, answer C.

\subsubsection{Annotation Guidelines for Math}

Please compare the two text data related to math and choose the one of higher
quality.

High-quality math data show significant mathematical intelligence and is
educational for math learners. Mathematical quality can be evaluated based on several
key aspects:

(1) Logical Structure: Content should demonstrate clear reasoning with properly
structured arguments, proofs and deductions, avoiding inconsistencies or
unjustified assumptions;

(2) Mathematical Rigor: Expressions should use precise and consistent notation,
terminology and symbols throughout, with all necessary steps clearly stated;

(3) Pedagogical Value: The content should be build systematically from
fundamentals to advanced ideas, including instructive examples that reinforce understanding;

(4) Conceptual Depth: Material should go beyond elementary arithmetic to explore
deeper mathematical concepts and problem-solving techniques, showing connections
between different ideas;

(5) Technical Accuracy: Content should be free of mathematical errors, misconceptions,
ambiguous notation, or incorrect terminology that could impede understanding.

High-quality mathematical content will excel in these areas while maintaining accessibility,
whereas lower-quality content may be lacking in one or more of these essential
aspects.

Choose the better one of A and B according to the above guidelines and your
preferences for mathematical quality. If the two texts are of similar level,
answer C.

\subsubsection{Annotation Guidelines for Logic}

Compare the following two texts, determine which one better requires and
promotes logical thinking by evaluating these three essential criteria:

1. Does understanding later content require careful reasoning from previous
information?

- Positive: Text that builds logical arguments progressively.

- Negative: Text that can be understood superficially without deeper thinking
Contextual Integration.

2. Does comprehension require connecting multiple pieces of evidence or ideas?

- Positive: Text with interconnected logical elements.

- Negative: Simple chronological narratives or disconnected descriptions
Structured Interpretation.

3. Can the content be understood through clear rational analysis?

- Positive: Text with well-defined logical relationships.

- Negative: Ambiguous literary expressions with multiple subjective
interpretations.

Choose the better one of A and B according to the above guidelines and your
preferences for logical quality. If the two texts are of similar level, answer C.

\section{Generated Prompts}

\subsection{Prompts Generated by TextGrad}
\label{sec:textgrad}

We show the prompts generated by TextGrad for the three domains in Section~\ref{sec:appendix_prompts}.
The quality criteria are in bold.

\begin{tcolorbox}
    [title = {Code}, breakable] \footnotesize \#\# Task Instruction$\backslash$nYou
    are tasked with performing a comprehensive comparison of the quality and structure
    of two Python code files. Evaluate them based on \textbf{readability, efficiency,
    adherence to Python coding standards (PEP 8), and maintainability}.
    Highlight strengths and weaknesses for each file and suggest specific
    improvements where necessary. $\backslash$n$\backslash$n\#\# Code File A$\backslash$n\{A\}$\backslash$n$\backslash$n\#\#
    Code File B$\backslash$n\{B\}
\end{tcolorbox}

\begin{tcolorbox}
    [title = {Math}, breakable] \footnotesize \#\# Compare the Mathematical Quality
    of Two Solutions$\backslash$nPlease evaluate the mathematical quality of the
    two provided solutions. Consider factors such as \textbf{correctness, clarity,
    logical reasoning, and mathematical rigor} in your assessment. Once you have
    thoroughly reviewed both solutions, choose "A" or "B" to identify the solution
    that exhibits superior mathematical quality.$\backslash$n$\backslash$n[A]$\backslash$n\{A\}$\backslash$n$\backslash$n[B]$\backslash$n\{B\$\}
\end{tcolorbox}

\begin{tcolorbox}
    [title = {Logic}, breakable] \footnotesize Assess the logical consistency
    between the two text pieces provided below. Identify which text is more \textbf{logically
    consistent}$\backslash$n of A, B, or if they are equally consistent. Clearly
    explain your reasoning behind the evaluation.$\backslash$n$\backslash$n[A]$\backslash$n\{A\}$\backslash$n[/A]$\backslash$n$\backslash$n[B]$\backslash$n\{B\}$\backslash$n[/B]
\end{tcolorbox}

\subsection{Prompts Generated by DRPO}
\label{sec:drpo}

\begin{tcolorbox}
    [title = {Code}, breakable] \footnotesize
    As an advanced language model, your role is to assist users by providing insightful, accurate, and engaging responses. Strive to be helpful, clear, and factual in your interactions, while offering depth and context where possible. Aim to provide comprehensive analyses and make clear recommendations or decisions when appropriate. Anticipate user needs and explore various perspectives to enrich your responses.$\backslash$n- You do not have access to the internet or real-time data, and you cannot perform physical actions. If a question is outside your capabilities, it is better to acknowledge this than to provide an unrelated response.$\backslash$n- When comparing two entities, such as Python code files, provide a thoughtful analysis based on the information available. If content is missing, suggest general criteria or considerations that could be used in a typical analysis.$\backslash$n- Avoid sounding robotic by using natural language and engaging with the user's query in a conversational manner.$\backslash$n- Ensure your responses are structured and easy to understand, even when dealing with limited information.$\backslash$n- Encourage the exploration of different perspectives or approaches when analyzing or comparing entities, to add depth to your responses.$\backslash$n- Strive to anticipate potential follow-up questions or areas of interest that the user might have, and address them proactively if possible.$\backslash$n- When faced with simple or limited inputs, consider broader contexts or related topics to enhance the depth and helpfulness of your response. Explore potential implications or applications of the topic to provide a richer response.$\backslash$n- Focus on enriching responses by connecting the topic to broader contexts or related fields, and consider potential implications or applications to provide a more comprehensive analysis.
\end{tcolorbox}

\section{Responsible NLP Research Statements}

We used generative AI to assist in this work. We used GitHub Copilot for short-form
input assistance when writing the code. We used ChatGPT and Claude for
paraphrasing and polishing the original content in the paper.

The datasets used in this work are publicly accessible. The usage of the Stack v2
is under Terms of Use for The Stack v2~\footnote{\url{https://huggingface.co/datasets/bigcode/the-stack-v2}}.
The usage of OpenWebMath is under ODC-By 1.0 license~\footnote{\url{https://opendatacommons.org/licenses/by/1-0/}}
and the CommonCrawl ToU~\footnote{\url{https://commoncrawl.org/terms-of-use/}}. The
usage of Zyda-2 is under the terms of Open Data Commons License~\footnote{\url{https://opendatacommons.org/licenses/by/1-0/}}.

We used \texttt{gpt-4o} for the experiments, which is under OpenAI's Terms of Use~\footnote{\url{https://openai.com/policies/terms-of-use/}}.
We used \texttt{Qwen2.5-72B-Instruct}, whose weight is distributed under Qwen
LICENSE AGREEMENT~\footnote{\url{https://huggingface.co/Qwen/Qwen2.5-72B-Instruct/blob/main/LICENSE}}.
We trained \texttt{Llama-3.2} respect to LLAMA 3.2 COMMUNITY LICENSE AGREEMENT~\footnote{\url{https://www.llama.com/llama3_2/license/}}.

\section{Prompts}
\label{sec:appendix_prompts}

\subsection{Prompts for Knowledge Base}

\begin{tcolorbox}
    [title = {Judge if a paper releases a dataset.}, breakable] \footnotesize
    There is a research paper about artificial intelligence.$\backslash$n$\backslash$nTitle:
    <TITLE>$\backslash$nAbstract: <ABSTRACT>$\backslash$n$\backslash$nInstruction:
    Does this paper propose a dataset? Return your answer in the following format:$\backslash$n$\backslash$n```
    json \{ "analysis": "Your analysis. For example, the main contribution of the
    paper.", "dataset": "The name of the dataset if it is proposed. Otherwise,
    answer 'N/A'.", "answer": "Yes/No/Unsure" \} ```
\end{tcolorbox}

\begin{tcolorbox}
    [title = {Extract quality criteria from papers.}, breakable] \footnotesize
    There is a research paper about artificial intelligence which proposed a new
    dataset named <DATASET\_NAME>.$\backslash$n$\backslash$n[BEGIN\_OF\_PAPER]$\backslash$n
    <PAPER\_CONTENT>$\backslash$n[END\_OF\_PAPER]$\backslash$n$\backslash$nI want
    to learn how to distinguish between data of high and low quality from the process
    of constructing the <DATASET\_NAME> dataset. Please conclude the criteria
    for determining data quality from the paper.$\backslash$n$\backslash$n- The
    criteria should be able to used to filter the data for the dataset.$\backslash$n
    - The criteria should be general enough to be applied to other datasets.$\backslash$n
    - If the paper proposed a data processing method, you should describe the criteria
    for the processed data which may be of higher quality.$\backslash$n - You
    should not just copy the criteria from the paper, but summarize them in your
    own words.$\backslash$n$\backslash$n```json \{ "name\_of\_the\_criterion": "description\_of\_the\_criterion",
    "name\_of\_another\_criterion": "description\_of\_another\_criterion", ...
    \} $\backslash$n$\backslash$nThe names of criteria should be a descriptive word.
    The descriptions should show what the criteria is about and how it can be
    used to determine if a data record should be included in the dataset. ```
\end{tcolorbox}

\begin{tcolorbox}
    [title = {Retrieve Code Criteria},breakable] \footnotesize \# Instruction$\backslash$nIs
    this criterion applicable for evaluating the quality of Python code?$\backslash$n$\backslash$n\#
    Criterion$\backslash$n<CRITERION>: <DESCRIPTION>$\backslash$n$\backslash$nYou
    should simply reply 'yes' or 'no'.
\end{tcolorbox}

\begin{tcolorbox}
    [title = {Retrieve Math Criteria},breakable] \footnotesize Is the following criterion
    applicable to measure the mathematical quality of text data?$\backslash$n$\backslash$n\#\#\#
    Criterion$\backslash$n*<CRITERION>*: <DESCRIPTION>$\backslash$n$\backslash$nYou
    should simply reply 'yes' or 'no'.
\end{tcolorbox}

\begin{tcolorbox}
    [title = {Retrieve Logic Criteria},breakable] \footnotesize \# Instruction$\backslash$nIs
    the following criterion applicable to evaluate the logical quality of text data?$\backslash$n$\backslash$n\#
    Criterion$\backslash$n<CRITERION>: <DESCRIPTION>$\backslash$n$\backslash$nYou
    should simply reply 'yes' or 'no'.
\end{tcolorbox}

\subsection{Domain Specific Prompts for Worker Agents}

\begin{tcolorbox}
    [title = {Pairwise Judgment for Code},breakable] \footnotesize \#\#
    Instruction$\backslash$nGiven criterion **{criterion}**, compare two Python code
    files and determine which one human annotators will consider to be of higher
    quality.$\backslash$n$\backslash$n\#\# A$\backslash$n\{A\}$\backslash$n$\backslash$n\#\#
    B$\backslash$n\{B\}$\backslash$n$\backslash$n\# Criterion$\backslash$n**\{criterion\}**:
    \{description\}
\end{tcolorbox}

\begin{tcolorbox}
    [title = {Pairwise Judgment for Math},breakable] \footnotesize \#\#
    Instruction$\backslash$nGiven criterion **\{criterion\}**, evaluate and determine
    which of the two text data is of higher quality in mathematics.$\backslash$n$\backslash$n[DATA\_A]$\backslash$n \{A\}$\backslash$n [/DATA\_A]$\backslash$n$\backslash$n[DATA\_B] $\backslash$n \{B\}$\backslash$n [/DATA\_B]$\backslash$n$\backslash$n\#
    Criterion$\backslash$n**\{criterion\}**: \{description\}
\end{tcolorbox}

\begin{tcolorbox}
    [title = {Pairwise Judgment for Logic},breakable] \footnotesize Which text piece
    of A and B is more logical based on **\{criterion\}**?$\backslash$n$\backslash$n\{criterion\}:
    \{description\}$\backslash$n$\backslash$n[A]$\backslash$n\{A\}$\backslash$n[/A]$\backslash$n$\backslash$n[B]$\backslash$n\{B\}$\backslash$n[/B]
\end{tcolorbox}

\subsection{Domain Specific Prompts for Manager Agents}

\begin{tcolorbox}
    [title = {Generate Initial Code Criteria},breakable] \footnotesize List and describe
    20 criteria on how human compare the overall quality of two Python code files.
\end{tcolorbox}

\begin{tcolorbox}
    [title = {Generate Initial Math Criteria},breakable] \footnotesize List and describe
    20 criteria on evaluating whether a text data is high quality math data.
\end{tcolorbox}

\begin{tcolorbox}
    [title = {Generate Initial Logic Criteria},breakable] \footnotesize List and
    describe 20 criteria to tell which is more logical of two text pieces.
\end{tcolorbox}

\subsection{General Prompts for \CritiQ Flow}

The full prompts of \CritiQ Flow are complex. We simply list the source code here.
Details can be checked in the \CritiQ repository.

\begin{lstlisting}
MANAGER_PROMPT_POSTFIX="""\n\nYour response should be in the following **json** format:
```json
{
"name_of_the_criterion": "Detailed description for the criterion such as what it is, how it can be evaluated, when it is applicable, and other relevant information. Be specific and detailed while keep concise.",
...
}
```"""
MANAGER_PROMPT_TEMPLATE="Give {n_criteria} criteria for evaluating data quality."
ACCURACY_PROMPT="The worker agents had evaluated data pairs aginst these criteria. The accuracy of each criterion is as follows:"
GOOD_CRITERIA_PROMPT_TEMPLATE="Accuracies of criteria {criteria} are over {threshold}. They are good criteria."
MID_CRITERIA_PROMPT_TEMPLATE="The accuracy of {criterion_name} is over {threshold_0} but less than {threshold_1}. It can be improved. Here is the raw description of the criterion:"
MID_CRITIQUE_PROMPT="This is an incorrect case:"
MID_A_PROMPT_TEMPLATE="[BEGIN_OF_A]\n{}\n[/END_OF_A]"
MID_B_PROMPT_TEMPLATE="[BEGIN_OF_B]\n{}\n[/END_OF_B]"
MID_HOWEVER_PROMPT_TEMPLATE="Against this criterion, the worker agent chose {wrong} as better, but the correct answer is {correct}. Here is how the worker agent thinks:\n\n{thought}"
MID_REFLECTION_PROMPT="""Please analyze this incorrect case together with the worker agent's thought. Based on your anaylsis, ples provide your critique for how to write a better description of this critierion to guide the worker make correct judgment or properly indicate inapplicable situations for this criterion.

Your response should be in the following **json** format:
{
"analysis": "Your analysis here.",
"critique": "How this criterion can be improved. Please just point out the key points in a few sentences."
}"""
MID_REFINE_PROMPT_TEMPLATE="There are the critiques for the wrong choices.\n\n{}\n\nBased on the above critiques, please improve the description for this criterion to make worker agents get higher accuracy. For exmaple, what it is, how it can be evaluated, when it is applicable, and other relevant information. Be specific and detailed while keep concise."
MID_FORMAT_PROMPT_TEMPLATE='Return the improved description in the following **json** format:\n\n{{"{criterion_name}": "The improved description"}}'
LOW_PROMPT_TEMPLATE="Criteria {criteria} have an accuracy of less than {threshold_0}. They should be removed from the criteria list. Please provide {num} new criteria. The new ones should not be duplicated with the above ones."
LOW_FORMAT_PROMPT="""Return the new criteria in the following **json** format:\n\n
```json
{
"your_better_criterion_here": "Detailed description for the criterion, including what it is, how it can be evaluated, when it is applicable, and other relevant information. Be specific and detailed while keep concise.",
...
}
```"""
PAIR_WORKER_PROMPT_POSTFIX="""
Your response should be in the following **JSON** format:
```json
{
"analysis_a": "Analyze A based on the given criterion.",
"analysis_b": "Analyze B based on the given criterion.",
"thought": "Compare A and B.",
"answer": "A / B / None"
}
```
Return None if any of the following conditions are met:
- The criterion is not applicable to this pair of data pieces.
- They are of the same quality.
- You are unsure.
"""
PAIR_WORKER_PROMPT="Which is better in the aspect of **{criterion}**?\n\n{criterion}: {description}\n\n[DATA_A]\n{A}\n[/DATA_A]\n\n[DATA_B]\n{B}\n[/DATA_B]"
CRITERION_FORMAT_TEMPLAT="\n\n[CRITERION]\nCriterion: {name}\n\nDescription: {desc}\n[/CRITERION]"
\end{lstlisting}

\section{Examples for Criteria Refinement}
\label{sec:appendix_ex_refinement}

In this section, we provide examples of the comparisons of the criterion before and after refinement, annotated by [BF] and [AF], respectively.

\subsection{Generated Criterion}
\begin{tcolorbox}
    [title = {[BF] Criterion \textbf{algorithm\_efficiency}},breakable] \footnotesize This criterion assesses the efficiency of the algorithm
    implemented in the code. It measures time complexity, space complexity, and
    overall performance under different input sizes. By comparing how well the code
    scales and performs, evaluators can determine which implementation is
    superior in terms of resource optimization. It is particularly applicable to
    competitive programming or performance-critical applications.
\end{tcolorbox}
\begin{tcolorbox}
    [title = {[AF] Criterion \textbf{algorithm\_efficiency}},breakable] \footnotesize The 'algorithm\_efficiency' criterion evaluates the computational
    performance of algorithms implemented within the code, focusing on
    measurable metrics such as time complexity, space complexity, and overall
    scalability. It assesses how well the code optimizes operations under varying
    input sizes and work complexity. This criterion is specifically applicable
    when the code implements clear and testable algorithms designed to solve
    computational or algorithmic tasks (e.g., sorting, searching, graph traversal).
    It is not concerned with the broader functionality, structure, or aesthetics
    of the code unless they directly impact efficiency. \par How to evaluate: Workers
    should focus on examining the implementation of algorithms for efficiency improvements
    based on established computational principles. Consider aspects like: 1. Does
    the algorithm exhibit better performance by reducing computational steps (time
    complexity)? 2. Does it use fewer resources, such as memory, to achieve its
    output (space complexity)? 3. How does the algorithm perform under different
    input sizes (scalability)? 4. Are there opportunities to optimize the
    algorithm further (e.g., using better data structures or reducing redundant operations)?\par
    When it is applicable: This criterion applies when both code files contain algorithms
    or computational content that can be directly analyzed for efficiency in terms
    of performance under realistic conditions. For example, files implementing tasks
    like sorting, searching, mathematical computations, or data processing
    should be evaluated under this criterion. \par When it is inapplicable: This
    criterion should be marked 'Not Applicable' when: 1. Neither file contains algorithmic
    or computational content (e.g., configuration files, import statements, or
    static setup code). 2. The files are of vastly different purposes or
    functionalities, making a direct comparison of algorithmic efficiency irrelevant.
    3. The code involves orchestration logic (e.g., API calls) rather than
    computational tasks. 4. The algorithms are incomplete, trivial, or lack sufficient
    context for meaningful evaluation (e.g., placeholders, stubs, or boilerplate
    code).\par Additional information: Workers should avoid judging efficiency
    based on perceived complexity or verbosity. Simpler code that efficiently achieves
    the desired result should be favored over unnecessarily complex solutions.
    Ensure all evaluations consider the problem the algorithm is designed to
    solve in its context. To assist with clarity, workers should consider
    explicitly marking the criterion inapplicable when one or both files fail to
    meet the stated requirements for algorithm efficiency evaluation. Clear examples:
    - Applicable: Comparing two sorting algorithms for time and space complexity.
    - Inapplicable: Comparing a configuration file to a data processing algorithm.
    - Inapplicable: Comparing an I/O-bound script with CPU-intensive code.\par This
    improved description ensures that workers consider the scope and context of 'algorithm\_efficiency'
    before making decisions, reducing ambiguity and inaccuracies.
\end{tcolorbox}

\subsection{Retrieved Criterion}
\begin{tcolorbox}
    [title = {[BF] Criterion \textbf{error\_analysis}},breakable] \footnotesize A thorough analysis of the reasoning flaws must be present.
    This includes understanding the causes of errors and what correct reasoning
    should have been applied at that point.
\end{tcolorbox}
\begin{tcolorbox}
    [title = {[AF] Criterion \textbf{error\_analysis}},breakable] \footnotesize Error analysis evaluates the quality and depth of reasoning related
    to identifying, diagnosing, handling, and mitigating potential or actual
    errors within the code. This includes examining error-handling mechanisms such
    as exception blocks, validation checks, logging, or any other explicit
    strategies to anticipate and address errors. Additionally, it considers the
    code's explanation or reasoning about errors, focusing on detail and thoroughness
    in addressing potential edge cases or failure points. To evaluate error
    analysis, workers should consider the following steps: (1) Identify the
    presence of error-handling logic or mechanisms in the code (e.g., try-except
    blocks, assertions, logging); (2) Assess whether the provided error-handling
    logic is appropriate for the scope and context of the code; (3) Pay
    attention to any accompanying comments or documentation explaining the
    approach to mitigating errors; and (4) Evaluate whether patterns of reasoning
    about errors are logical and well-structured, including how edge cases are
    anticipated.\par This criterion is applicable only to code that contains
    logical processes, algorithms, or decision-making components where errors
    are likely to occur and need to be reasoned about or handled. It should be marked
    inapplicable for code that lacks relevant error-handling context, such as
    configuration files, boilerplate code, or import-only scripts. In cases where
    both pieces of code lack any mention or handling of errors, the criterion
    should also be deemed inapplicable, and no preference should be made.\par Key
    aspects to avoid include judging the code based on its overall complexity,
    functionality, or modularity unless they directly affect error analysis.
    Highlighting superficial error handling or assuming error-free code does not
    inherently satisfy this criterion. Workers should focus on explicit reasoning
    about errors and how the code mitigates or avoids potential failures.
    Concrete examples of good error analysis include thorough exception handling
    with explanations, detailed error logging, validations targeting specific
    failure scenarios, and robust test cases explicitly aimed at uncovering edge
    cases or logical flaws.
\end{tcolorbox}

\subsection{Refined Criterion}
\begin{tcolorbox}
    [title = {[BF] Criterion \textbf{commented\_context}},breakable] \footnotesize The 'commented\_context' criterion evaluates the presence,
    relevance, and quality of comments or documentation within a code file,
    ensuring they enhance understanding of the code's purpose, functionality,
    and any non-obvious logic. Comments should provide meaningful insights about
    the code's intent, clarify complex or non-intuitive sections, and offer context,
    such as explaining critical operations or unusual design decisions. This criterion
    does not favor the mere presence of comments or their verbosity but instead focuses
    on their necessity and utility in aiding comprehension. \par Approach for
    evaluation: Workers should assess whether comments are directly relevant to specific
    parts of the code and whether they provide significant contextual value to understanding
    its intent and usage. For instance, comments explaining business logic,
    algorithmic choices, or intricate areas of code are highly valuable.
    Irrelevant, redundant, or excessively verbose comments that do not add
    clarity should not be positively weighted. Self-documenting code, where the
    use of clear variable/function names and logical structure makes it
    inherently understandable, should not be penalized for a lack of comments.
    \par Applicability: This criterion is most relevant when comparing code that
    requires additional explanation due to complexity or specialized logic. It
    is less applicable or should be marked inapplicable when both files contain
    minimal or no comments, but their code is simple and self-explanatory.
    Examples include boilerplate files, import-only files, or scripts so
    straightforward that no additional context is needed. \par Additional considerations:
    Workers should not rely on style or verbosity as sole indicators of quality.
    Comments that are overly generic (e.g., 'This is a for loop') or unrelated (e.g.,
    boilerplate licensing information) should not factor into the evaluation.
    When both files feature sufficient documentation for their respective levels
    of complexity, preference should be given to concise, context-rich comments
    over verbose or unnecessary ones. If both files lack meaningful comments and
    are equally understandable without additional documentation, this criterion
    may not provide a basis for comparison.
\end{tcolorbox}
\begin{tcolorbox}
    [title = {[AF] Criterion \textbf{commented\_context}},breakable] \footnotesize The 'commented\_context' criterion evaluates the presence,
    relevance, and necessity of comments or documentation within a code file.
    Comments should meaningfully enhance understanding by providing critical
    context, explaining complex logic, or clarifying non-obvious design
    decisions. The value of comments should be judged by their ability to aid comprehension,
    rather than their quantity or verbosity. High-quality comments are concise,
    appropriately placed, and directly related to the code's purpose and functionality.
    For example, comments explaining intricate algorithms, decision-making
    processes, or domain-specific details are valuable, whereas redundant, trivial,
    or boilerplate comments (e.g., licensing headers, generic statements like 'this
    is a loop') are not.\par Evaluation Steps: 1. Assess whether the file
    contains comments, and if present, determine whether they address essential
    aspects of the code's logic, design, or purpose. 2. Focus on relevance:
    Identify whether the comments clarify concepts that are not immediately
    understandable from the code structure itself. 3. Consider necessity:
    Evaluate if the complexity of the code requires additional explanation, or if
    the code is inherently self-explanatory (e.g., simple utility scripts or
    well-named variables/functions). 4. Judge quality: Favor concise, meaningful
    comments over verbose, generic, or redundant ones. 5. Evaluate whether comments
    contribute to maintainability by providing future developers with clear insights
    into the code's intent or potential edge cases.\par Applicability: - This
    criterion is applicable when the code includes non-obvious logic, intricate design,
    or contextual details that are essential for understanding. For example, it
    applies to files with algorithms, configuration settings, or any code where
    additional clarification adds significant value. - It is not applicable for files
    containing minimal or self-explanatory code, such as import statements, trivial
    scripts, or boilerplate content, where comments are unnecessary. - When
    comparing two files, if both lack comments but are sufficiently self-documenting,
    this criterion should be marked as inapplicable rather than favoring one
    file over the other based on the absence of comments.\par Additional Notes: -
    Avoid penalizing files that are simple and naturally clear without requiring
    comments. Instead, prioritize whether the comments add actual value relative
    to the code’s complexity. - Clear examples should be provided to illustrate
    appropriate use, such as comments that explain unexpected behavior or unconventional
    approaches, versus meaningless or excessive commentary that does not enhance
    comprehension. - Do not elevate files with verbose or irrelevant comments over
    those with concise, targeted, and effective comments. Focus on substance, not
    volume. - Metadata comments, like licensing information, may be required for
    compliance but should not be counted as contributing to 'commented context'
    unless they add value to the understanding of the code.\par In summary, this
    criterion focuses on whether comments are necessary, relevant, and useful in
    providing additional context or understanding. It recognizes that not all
    code requires extensive commenting and explicitly allows for marking the
    criterion as 'Not Applicable' in cases of minimalistic, self-explanatory, or
    trivial files.
\end{tcolorbox}